# A Novel Motion Detection Method Resistant to Severe Illumination Changes

Sahar Yousefi[1,2] Mohammad T. Manzuri Shalmani[1] Jeremy Lin[3] Marius Staring[2,4]
Sharif University of Technology, Tehran, Iran
Leiden University Medical Center, Leiden, The Netherlands
PJM Interconnection, Audubon, PA 19403, USA
Delft University of Technology, Delft, The Netherlands

*Abstract*— **Recently, there has been a considerable attention given to the motion detection problem due to the explosive growth of its applications in video analysis and surveillance systems. While the previous approaches can produce good results, an accurate detection of motion remains a challenging task due to the difficulties raised by illumination variations, occlusion, camouflage, burst physical motion, dynamic texture, and environmental changes such as those on climate changes, sunlight changes during a day, etc. In this paper, we propose a novel per-pixel motion descriptor for both motion detection and dynamic texture segmentation which outperforms the current methods in the literature particularly in severe scenarios. The proposed descriptor is based on two complementary three-dimensional-discrete wavelet transform (3D-DWT) and three-dimensional wavelet leader. In this approach, a feature vector is extracted for each pixel by applying a novel three dimensional wavelet-based motion descriptor. Then, the extracted features are clustered by a clustering method such as well-known *k*-means algorithm or Gaussian Mixture Model (GMM). The experimental results demonstrate the effectiveness of our proposed method compared to the other motion detection approaches from the literature. The application of the proposed method and additional experimental results for the different datasets are available at (http://dspl.ce.sharif.edu/motiondetector.html).**

**Keywords- Motion detection, Dynamic texture, 3D-discrete Wavelet Transform, Wavelet leader.**

## I. INTRODUCTION

Over the past decade, the problem of motion detection is attracting more attention due to its wide range of applications in video surveillance, natural disaster investigation systems, and other areas. For this purpose, a wide variety of approaches to solve this problem has been proposed in the literature [1-12]. The proposed approaches in the literature can be divided into two categories: 1) spatial domain methods, and 2) frequency domain methods.

In spatial domain approaches, spatiotemporal descriptors are often employed in order to model the motion patterns by ignoring the holistic motion patterns. St-Charles et al. [12] proposed Self–Balanced SENsitivity SEgmenter (SuBSENSE). In this method, the authors used a spatiotemporal local binary similarity pattern (LBSP) for characterizing the pixel representations in a nonparametric paradigm, which is tuned by pixel level feedback loops. In other words, this method tries to model the background using the pixel level feedback loops. Therefore a background tuning process is done for modeling the background. LBSP defines for each pixel $p$ a neighbor set $N(p)$ on the frames, and then assigns a binary pattern to the pixel $p$ based on the differences of gray-levels between the neighbor pixels, $q \in N(p)$, and $p$. If the illumination variation were non-uniform (i.e. the gray-level of a sub-set of neighbors changed), the binary pattern will be changed. Therefore, LBSP is not robust under non-uniform illumination variation. Also, in order to regularize the process and eliminate the salt-pepper noise, SuBSENSE uses the morphological operations and median filter. These operations cause to eliminate far and tiny moving objects. Moreover, the background tuning process is usually slow and makes it difficult to adapt to the sudden illumination variations and burst motions [9]. Bianco et al. [10] exploited Genetic Programming and combined the state-of-art of the motion detection approaches in order to get the best solution. This method suffers from heavy computation burden. Moreover, there is no guarantee of finding the global maxima.

On the contrary, frequency domain approaches can consider holistic motion pattern information [13]. In [14] and [15], the two-dimensional discrete wavelet transform (2D-DWT) is used for moving object detection. These methods compare 2D-DWT of the current frame with the previous frames to detect a motion. For compression, a threshold value is defined. These methods do not consider the intrinsic temporal dimension in wavelet coefficients computation and hence the results are sensitive to the predefined threshold.

While good results can be achieved in the approaches mentioned earlier, accurate motion detection remains a challenging task due to the difficulties raised by illumination variations, occlusion, camouflage, burst physical motion, dynamic texture, and environmental changes such as those on climate changes, sunlight changes during a day, etc. In this paper, we propose a novel motion detection method which defines a pixel-based features descriptor based on three-dimensional discrete wavelet transform (3D-DWT) representations which can overcome the stated difficulties. By applying the well-known 2D-wavelet transform on an image, the approximation coefficient and the detail coefficients for three levels are computed for each of the three decomposition directions (horizontal, vertical and diagonal detail coefficients). By applying 3D-wavelet transform on a volume, the approximation coefficient and seven detail coefficients in seven different directions are computed. Therefore, frequency domain method can consider holistic motion pattern information. These coefficients in special directions can model the motion pattern across the time (the consequence frames). In this paper, we propose a new per-pixel motion descriptor based on 3D-DWT, which can obtain some feature vectors for modeling the motion pattern.

There are some potential challenges in change detection algorithms. Camouflage identification is one of the challenging issues in motion detection. Camouflage is a situation for which motion detection is difficult because of the similarity between the color of dynamic objects and the background [16]. In [12] and [17] spatiotemporal binary features and color information are used for detecting the camouflaged foreground objects. Ramiez et al. [16] proposed a thresholding approach which tunes the values of the thresholds based on the analysis of the global Hue histogram of the frames. Due to its global character, this method cannot account for the frames which a color predominates on the scene locally. In our proposed method, which uses wavelet-based spatial frequency descriptors, we can achieve a high degree of insensitivity to camouflaged foreground objects.

Burst physical motion detection, which was not considered seriously, is another challenging issue in motion detection problems. Moving escalators, swirling wheels, fans, swinging foliage are some of the examples of the burst physical motion. With regards to the background tuning approaches, such as [11, 12, 18], which update the background regions using the history of the pixels with a fixed learning rate, they are sensitive to burst motions [9]. Liang et al. proposed a frequency and speed adaptive background model for this purpose [3, 9]. In our proposed approach, the burst physical motion detection is handled due to its intrinsic frequency-based motion pattern descriptors.

The dynamic texture issue is the last challenging issue which we consider. To our knowledge, none of the present approaches consider the dynamic textures as the moving regions. Due to the wavelet coefficient-based descriptors which describe motion patterns by their motion behavior, our proposed approach is able to tolerate the dynamic textures.

In the proposed approach, a representation of spatial frequency motion pattern based on three dimensional low-pass and high-pass wavelet coefficient as well as wavelet leaders is provided first. Wavelet leaders overcome the problem of a large number of close-to-zero wavelet coefficients [19]. Then a set of pixel-based features over these wavelet coefficients are computed. After feature extraction, the vectors are clustered by using well-known *k*-means algorithm.

The three main contributions of this paper are:

(1) A robust and effective approach for motion detection which outperforms the previous methods for videos with difficult environment like changing weather conditions, illumination variations, camouflaged foreground objects, and others;

(2) A novel method which detects the background motions based on their motion patterns such as burst physical motions, dynamic textures;

(3) A spatial frequency per-pixel basis feature extraction method based on high-pass/low-pass wavelet coefficients and wavelet leader for achieving the proper motion detection results and preventing the jagged boundaries.

The remainder of this paper is organized as follows: Section II describes an overview of the proposed method while Section III explains the three-dimensional wavelet transform and wavelet leader. Section IV describes the proposed wavelet-based motion descriptors. The efficiency evaluation of the proposed model through various video sequences in comparison with the previous approaches is presented in Section V. Finally, Section VI concludes.

## II. OVERVIEW OF THE PROPOSED METHOD

In this section, we provide an overview of the proposed approach. Figure 1 illustrates the flowchart of the proposed method. Also, the implementation of the proposed method, named *3D-DWT motion detector (3D-DWT_MD)* tool, is freely available at our Motion Detection Webpage: http://dspl.ce.sharif.edu/projects/MotionDetector/MotionDetector.rar.

### A. Deinterlacing

Deinterlacing is a process in which the video sequences are converted into progressive sequences. The datasets which are used in this paper, used the spatiotemporal median filter for de-interlacing the videos. The spatiotemporal median filter is a non-linear filter which is obtained by extending the spatial median filter to spatiotemporal neighbors [20, 21].

### B. 3D-Discrete Wavelet Transform (3D-DWT)

To consider the motion and appearance of the video sequences, we use three-dimensional discrete wavelet transform (3D-DWT) method. As mentioned before, the 2D-wavelet transform computes the approximation coefficient and the three levels detail coefficients contain horizontal, vertical and diagonal detail coefficients for a two dimensional data (e.g. an image). Applying the 3D-wavelet transform on a three dimensional data (i.e. a volume) computes the approximation coefficient and seven detail coefficients in seven different directions. These coefficients in special directions can model the motion pattern across time in different directions. Hence, by using frequency domain, we can consider the holistic motion pattern information. In this paper, we propose a new pixel-based motion descriptor based on 3D-wavelet coefficients, which models the motion pattern in three dimensional spaces. 3D-DWT can be coupled into two steps: 2D-spatial DWT and 2D-temporal DWT. In other words, the two-dimensional spatial and temporal transforms are done separately [22]. As will be described in the following section, the spatial DWT considers the holistic motion pattern information in space and the temporal DWT considers the holistic motion pattern information across time. In our proposed method, 3D-DWT is applied to the volumes (cubic patches). After each level of performing 3D-DWT on a volume, eight sub-cubic patches are created. This process is shown for one level in figure 1.

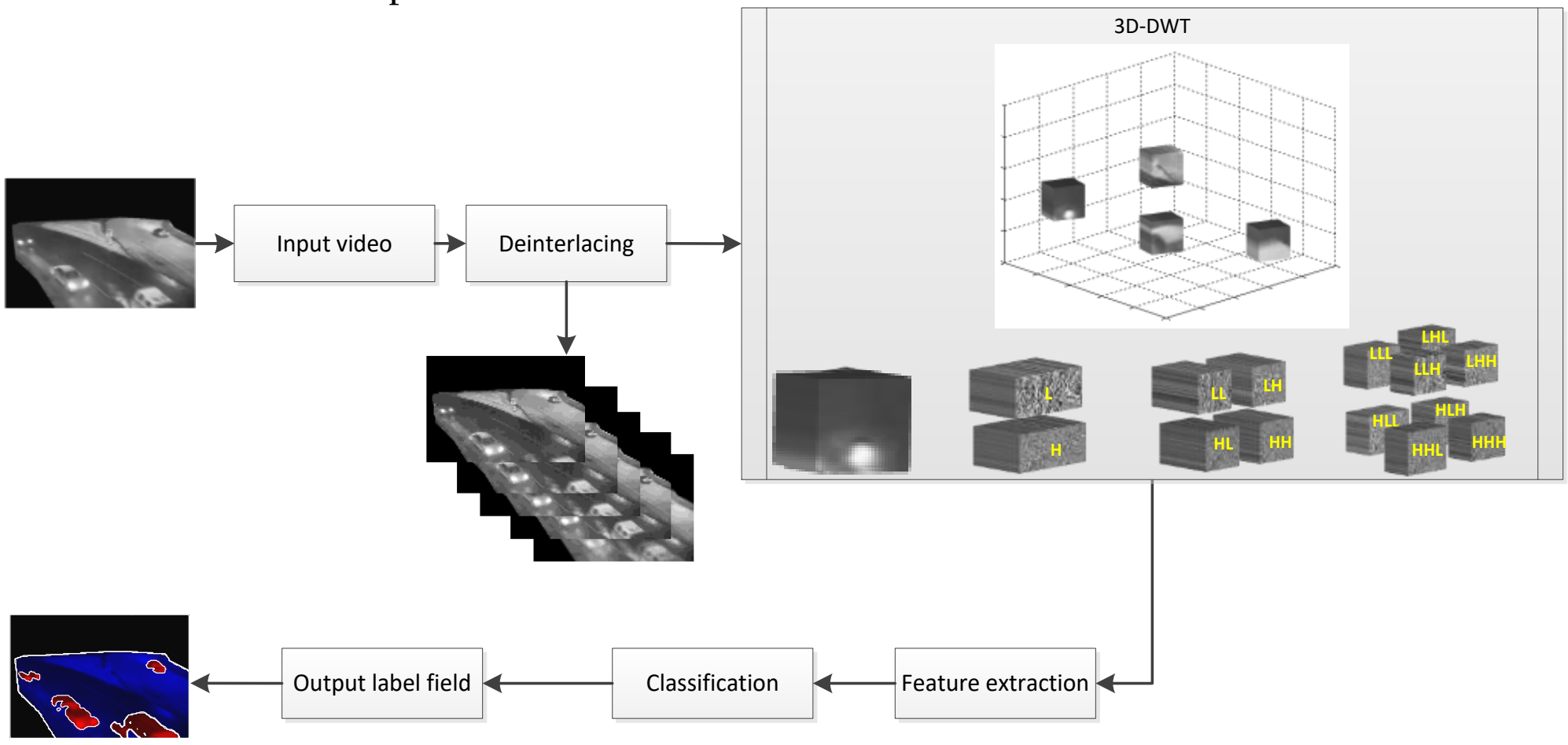

Figure 1. Flowchart of the proposed method

### C. *Feature Extraction*

In order to obtain the proper results and prevent the jagged boundaries, the feature vectors are described by a pixel-based process over the outputs obtained in the previous step. Zhang et al. proposed some feature vectors on 2D-DWT [23]. In our work, we propose eight feature vectors on 3D-DWT decomposition coefficients and 3D-wavelet leader which will be described in the following sections.

### D. *Classification*

After feature extraction, the feature vectors are classified into two different classes: *motion* vs. *zero-motion*. In *3D-DWT_MD* we contrived two different approaches, which contain *k-means* and *Gaussian Mixture Model (GMM)*, for this goal. In this paper, the results are reported using only *k-means* classification approach.

## III. Wavelet and Wavelet Leader

The 2D-DWT are widely used for moving texture detection such as smoke detection [24], fire detection [25], etc. Demonceaux et al. [26] proposed the combination of 2D-DWT and hierarchal Markov random fields for motion detection. In order to overcome the problem of temporal aliasing, an estimation of dominant motion on several image resolutions is obtained. In this work, the segmentation results exhibit obvious jaggedness of the boundaries. In order to prevent jagged boundaries, we propose a three-dimensional wavelet coefficient feature-based approach for motion detection.

In the one-dimensional wavelet transform, a function $f(t)$ can be analyzed by:

$$f(t)=\sum_{\tau,s}\Psi_t(\tau,s)\varphi_t(\tau,s), \qquad (1)$$

where $\Psi_t(\tau,s)$s are the wavelet coefficients estimated by the inner product via,

$$\Psi_t(\tau,s)=\int_{-\infty}^{\infty} f(x)\varphi_t(\tau,s)dt\ . \qquad (2)$$

Each of the wavelet coefficients $\Psi_t(\tau,s)$ represents the resemblance of the function $f(t)$ to the wavelet bases $\varphi_t(\tau,s)$ at a specific translation $\tau$ and scale $s$. In this paper, we use Coiflet-like nearly symmetric orthogonal wavelet bases with magnitude and group delay flatness specification which has been proposed in [27]. Using the multi-scale wavelet decomposition scheme, a hierarchy of localized sub-functions at different spatial frequencies can be found.

As mentioned before, a 3D-DWT can be coupled into two steps: 2D-spatial DWT and 2D-temporal DWT. As also previously mentioned, considering 3D-DWT into spatial and temporal dimensions leads to gain holistic motion pattern information through space and time. Suppose we have volume $V$, the 3D-DWT decomposes the volume $V$ into (1) one low-frequency $w_{LLL}^{(S)}$, (2) seven strict high-frequency channels $\tilde{w}^{(S)}$, (3) and multiple non-strict high-frequency channels $w_{o,l}^{(s)}$ for $s\in\{1,\ldots,S-1\}$, which $s$ is the scale set, $S$ is the coarsest scale, $o\in O$ is the orientation set, where $O=\{vertical, horizontal, diagonal\}$, and $l\in\Gamma$ is the level set, where $\Gamma=\{up, down\}$.

Decomposing a volume $\Lambda$, into the wavelet coefficient set is a recursive process function $\Phi^{(s)}(\Lambda)$ for $s=S,\ldots,1$ in which,

$$\Phi^{(s)}(\Lambda)=\begin{cases}\psi^{(s)}(\Lambda) & s=1\\ \left\{\psi^{(s)}(\Lambda)-w^{(s)}_{LLL}(\Lambda)\right\}\cup\Phi^{(s-1)}\left(w^{(s)}_{LLL}(\Lambda)\right) & s>1\end{cases}, \quad (3)$$

and

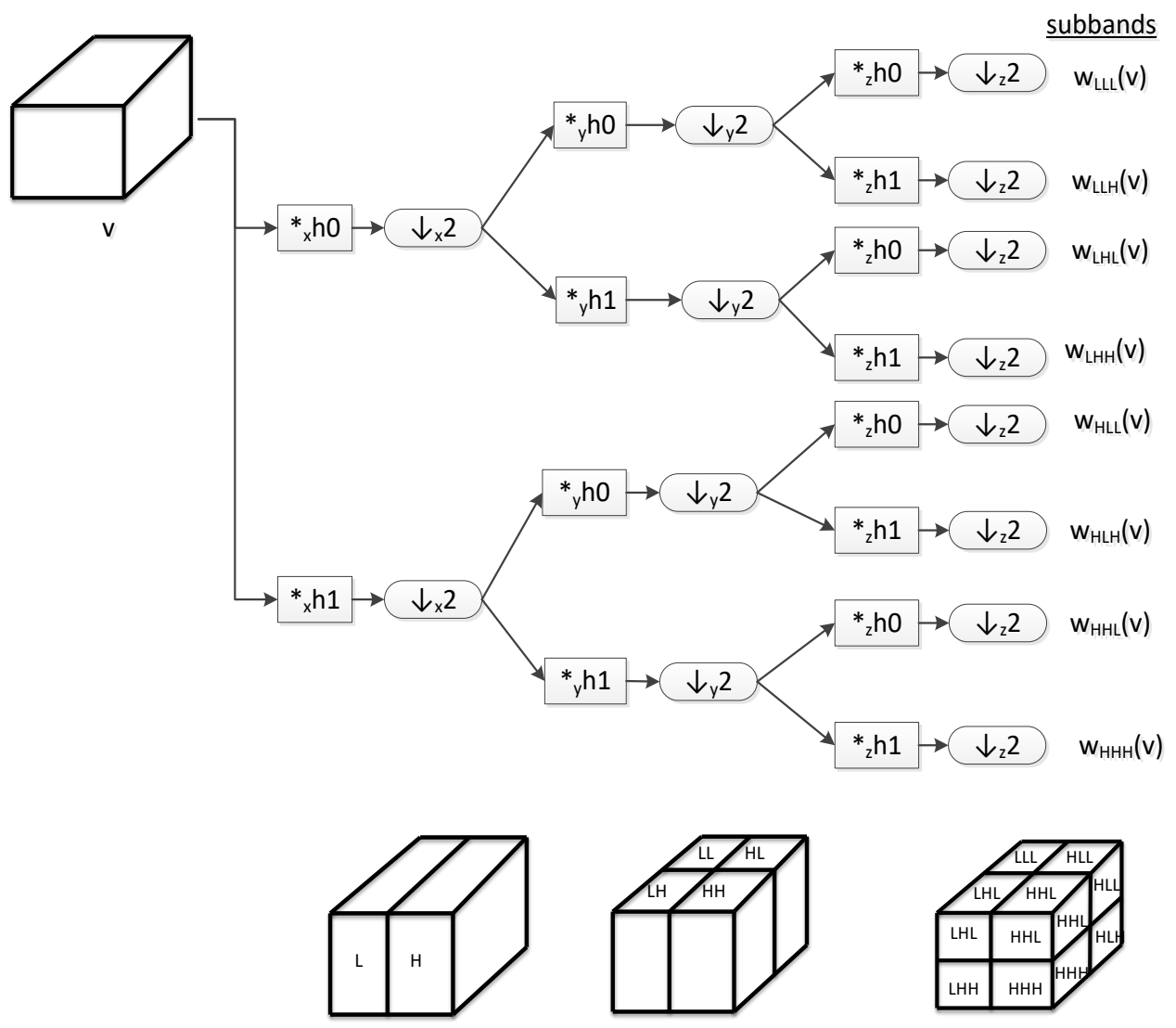

Figure 2: Three Dimensional Discrete Wavelet filters bank

$$\psi^{(s)}(\Lambda)=\begin{Bmatrix}w^{(s)}_{LLL}(\Lambda), w^{(s)}_{LLH}(\Lambda), w^{(s)}_{LHL}(\Lambda), w^{(s)}_{LHH}(\Lambda),\\ w^{(s)}_{HLL}(\Lambda), w^{(s)}_{HLH}(\Lambda), w^{(s)}_{HHL}(\Lambda), w^{(s)}_{HHH}(\Lambda)\end{Bmatrix} \quad (4)$$

Regarding filter concepts, Figure 2 illustrates the 3D-DWT filter banks for $\Phi^{(1)}(V)$. The figure indicates the sub-volumes of each filter for one scale and a volume *V*.

In order to improve the robustness of descriptors of the wavelet coefficients, we use Wavelet leaders [28]. Wavelet leaders are another wavelet-based measurements which are defined in [28] for the first time. In the literature, wavelet leaders are used for various applications [13, 29-32]. Chen et al. used the wavelet leader pyramids for image quality assessment [30]. Pustelnik et al. proposed a combination of wavelet leaders and proximal minimization in order to segment textures in images [31, 32]. Ji et al. used wavelet leaders for dynamic texture classification [13].

Wavelet leaders are defined as the maximum magnitude of all the wavelet coefficients for local spatial neighborhood and scale neighborhood. As mentioned before, wavelet coefficients compute a large number of close-to-zero values. In this paper, in order to obviate the problems raised by wavelet coefficients, we propose three-dimensional wavelet leader pyramids. The wavelet leaders for a volume $\Lambda_p$, surrounding the pixel $p$, are defined as:

$$w^{(s)}_{Leader}(\Lambda_p)=\max_{o\in O}\max_{l\in\Gamma}\left\{\left|w^{(s)}_{o,l}\right|,\left|\tilde{w}^{(s)}\right|\right\}\ for\ 1\le s\le S-1 . \quad (5)$$

Figure 3 represents $\Lambda_p$ with size $(8\times 8\times 8)$. In this figure, the volume is defined by selecting $(8\times 8)$ neighborhood windows on a sequence of frames with the length 8, the red site is $p$ and the other sites compose $\Lambda_p$. Figure 4 illustrates the three dimensional wavelet coefficients and the wavelet leaders of a volume for three scales based on equations (3) and (4). In this Figure the volumes in (3) and (4) are removed for brevity.

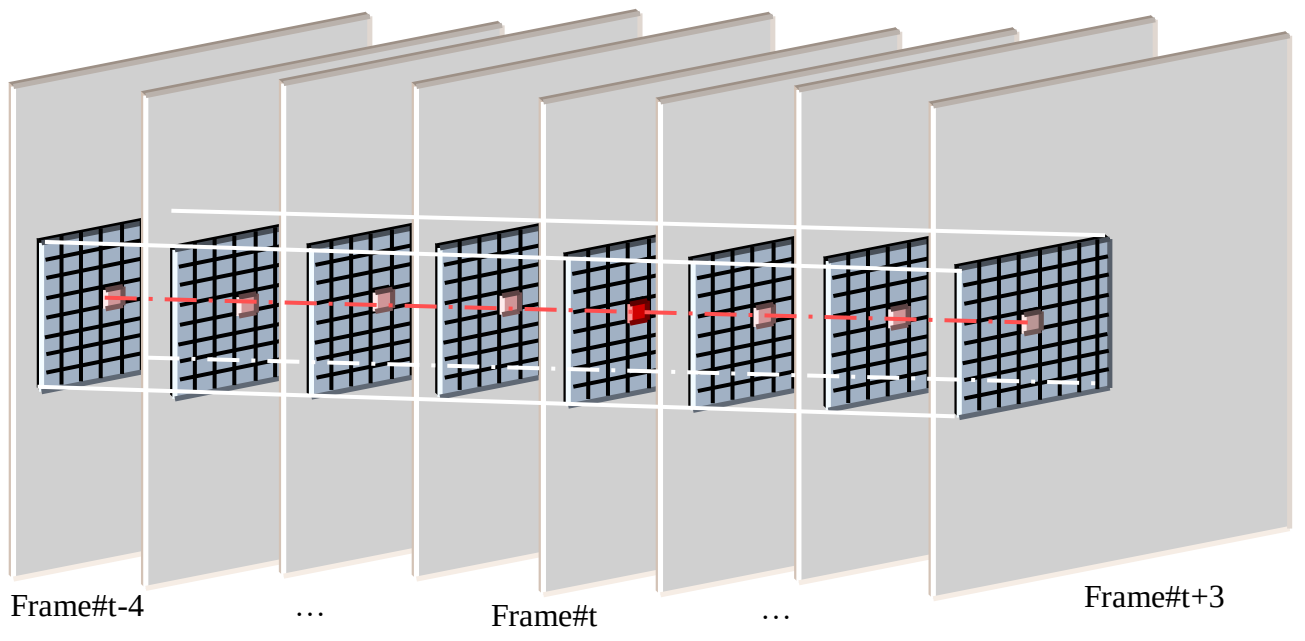

Figure 3: $\Lambda_p$ with size 8×8×8 on a sequence of frames, the red site is *p* and the blue sites are the neighborhood set $\Lambda_p$

## IV. Wavelet-Based Motion Descriptors

Using 3D-wavelet coefficients, we can model the holistic motion patterns in different directions. By applying 3D-wavelet transform on a volume or a cubic patch, the approximation coefficient and seven detail coefficients in seven different directions in a 3D space are computed. The high-pass wavelet coefficients can model the motion patterns over a sequence of frames. In this section, we propose a novel per-pixel motion descriptor based on 3D-DWT, which can model the motion pattern. Pixel-based feature vectors are helpful for achieving proper results and prevent jagged boundaries.

The descriptor vectors are defined as:

$$f(p)=\left(\sum_{s=1}^{S}\left(\sqrt{\sum_{r'\in\Lambda_p}\left(w^{(s)}_{\eta}\left(r^{(s)}\right)\right)^{2}_{r'}}\right)\right)\Big/\sum_{s=1}^{S}s , \quad (6)$$

where $p$ is the central pixel on a neighborhood cube $\Lambda_p$, $r'$ shows the neighbors of $p$ in $\Lambda_p$, and $r^{(s)}$ is defined as below,

$$r^{(s)}=\begin{cases}\Lambda_p & s=S\\ w^{(s+1)}_{\eta}\left(r^{(s+1)}\right) & 1\le s<S\end{cases}, \quad (7)$$

in which $S$ is the coarsest scale, $w^{(s)}_{\eta}$ demonstrates the wavelet sub-bands of the $s^{th}$ scale where

$$\eta\in\{LLL, LLH, LHL, LHH, HLL, HLH, HHL, HHH, Leader\}. \quad (8)$$

In equation (6), square root of sum of squares, i.e.

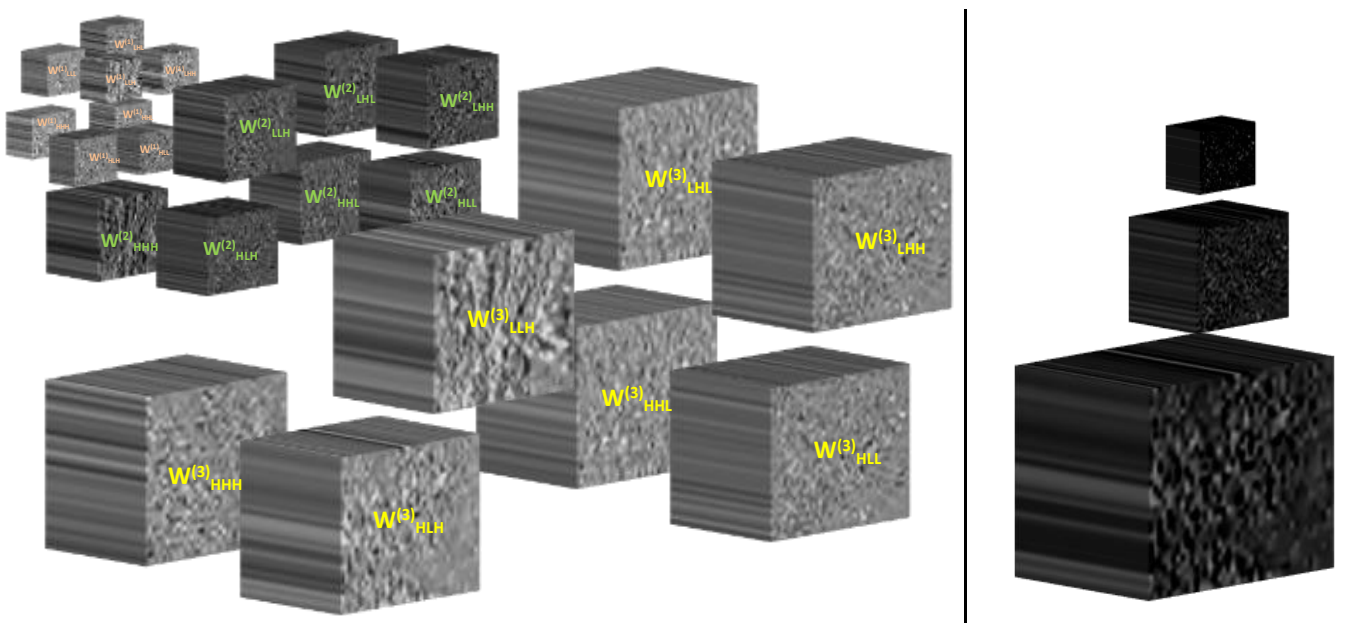

(a) (b)

Figure 4: (a) wavelet coefficients for $\Phi^{(3)}(\Lambda)$, (b) wavelet leaders for $\Phi^{(3)}(\Lambda)$

$\sqrt{\sum_{r'\in\Lambda_p}\left(w_\eta^{(s)}\left(r^{(s)}\right)\right)_{r'}^2}$, computes the Euclidean norm on a cubic-patch-size-dimensional space for the $s^{th}$ scale level of the $\eta$ wavelet sub-bands, which gives the ordinary distance from an origin to the patch vector. Considering averaging over multiple scale levels in equation (6) leads to apply the more significant wavelet components in finer scales which consequently causes noise reduction. In this paper, in order to compute the feature descriptors, we use $S$ scales for the volumes with size ($2^S \times 2^S \times 2^S$). Figure 5 illustrates the feature vectors extracted from the wavelet coefficients and the wavelet leaders for a set of consecutive frames, using cubic patches with size $(4\times4\times4)$ and $(S=2)$. As results indicate, the feature vectors extracted from high-pass wavelet coefficients contain $w_{LHH}^{(s)}$, $w_{HLH}^{(s)}$, and $w_{LLH}^{(s)}$, illustrate significant motion patterns. Hence, we use these feature vectors and the acquired wavelet leader from these features for modeling the motion patterns in three dimensional space.

## V. EXPERIMENTAL RESULTS

In this section, in order to evaluate the performance of the proposed approach, the results obtained are reported both qualitatively and quantitatively on video datasets including a variety of indoor and outdoor environments. The qualitative results are compared with various methods including CP3-online [9], IUTIS-3 [10], SUBSENSE [12], AAPSA [16], and MDT [33]. The quantitative results are reported by comparing them with the results of several unsupervised approaches including CP3-online [9], IUTIS-3 [10], SUBSENSE [12], AAPSA [16], C-EFIC [34], GMM|Zivkovic[35], CwisarDH [4], SOBS_CF [2], and AMBER [36]. We evaluate the proposed method on four datasets:

1. In order to evaluate our method for motion detection under various light conditions, we use CD.net2014 dataset (available at [www.changedetection.net](www.changedetection.net)). This dataset provides a realistic, camera-captured (no CGI), diverse set of videos which contains several video categories with 4 to 6 video sequences in each category [37]. Furthermore, each category is accompanied by accurate ground-truth segmentation and annotation of change or motion areas for each video frame. In order to compare our method with the previous methods, we obtained the results of the previous methods which were reported and maintained on our Motion Detection Webpage. For this goal, we compare the proposed method with the previously reported unsupervised methods which are published in the first-tier conferences and journals in the recent years. Whereas we do not consider the camera motion such as camera shaking or irregular camera motion such as camera jitter, we report the experimental results for the video sequences which are captured by a static camera. For this purpose, we examine the proposed method for three different categories of this dataset, which contains *Nightvideos*, *Thermal*, and *intermittentObjectMotion*.

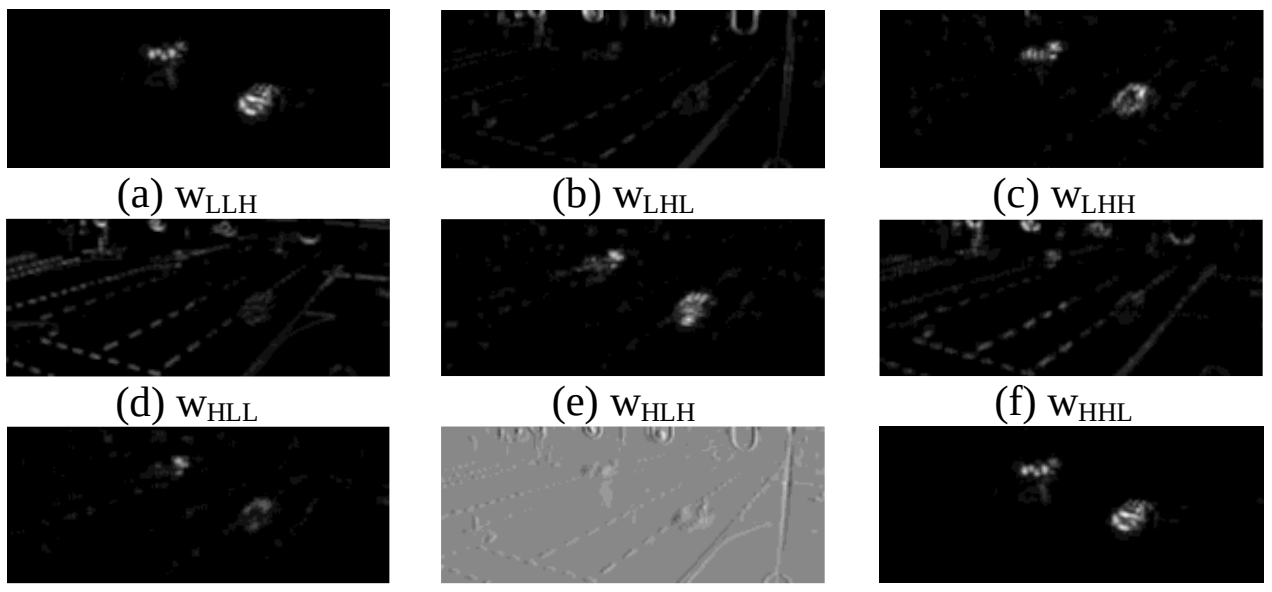

(a) $w_{LLH}$ (b) $w_{LHL}$ (c) $w_{LHH}$ (d) $w_{HLL}$ (e) $w_{HLH}$ (f) $w_{HHL}$ (g) $w_{HHH}$ (h) $w_{LLL}$ (i) Leader

Figure 5: (a-g) Features extracted from high-pass wavelet coefficients, (h) Feature extracted from low-pass wavelet coefficients, (i) wavelet leader extracted from high-pass wavelet coefficients (LLH, LHH, HLH)

2. For evaluating the proposed method for burst detection, Wallflower dataset [38] is used which contains burst motion video sequences. For this purpose, we use two different scenarios of this dataset - Camouflage, and Waving Trees.

3. As mentioned before, dynamic texture detection can be considered as motion detection problem. In order to examine the proposed method for dynamic texture detection, Dyntex dataset [39] is used. Dyntex is a comprehensive database of dynamic textures providing a large and diverse database of high-quality dynamic textures, which have been de-interlaced with a spatiotemporal median filter. The dynamic texture sequences have been acquired using a SONY 3 CCD Camera and a tripod. The sequences are recorded in PAL format (720 × 576).

4. UCSD pedestrian dataset [33] is another dataset which is used for motion detection evaluation. The dataset contains video of pedestrians on UCSD walkways, taken from a stationary camera with two different viewpoints. In this paper, we compare the results of the proposed method to the results of MDT [33] (available at [http://visal.cs.cityu.edu.hk/](http://visal.cs.cityu.edu.hk/) ).

### A. *Qualitative Comparison*

In this section, the qualitative comparison of our proposed method with various methods is considered. As mentioned before, our proposed method is compared to various approaches including CP3-online [9], IUTIS-3 [10], SUBSENSE [12], AAPSA [16], and MDT [33].

From the viewpoint of the occlusion, Figure 6 illustrates the comparison of the moving object segmentation between the proposed method and the previous approaches, including CP3-online [9], IUTIS-3 [10], and SUBSENSE [12], for two frames of the *winterStreet* sequence. In the figure, the red segments are the gray mask extracted from the ground truth, the blue segments illustrate the moving regions, the green segments represent the static regions, and the yellow circles highlight the differences of the segmentation results. As can be seen in the results, our proposed method overcomes the occlusion problem.

Figure 7 illustrates a more qualitative comparison of our proposed method with the aforementioned approaches, for various frames of *winterStreet* sequence of the CD.net 2014. As shown in the results, despite severe environmental conditions raised by video acquisition and car light at night, our proposed method can deal with the occlusion properly, unlike the other methods.

Another qualitative comparison, for various frames of *streetCornerAtNight* sequence of the CD.net 2014 dataset, is indicated in Figure 8. The results of our proposed method are compared with CP3-online [9] and SUBSENSE [12]. As shown in the results, our proposed method can overcome the motion detection problem at night light properly.

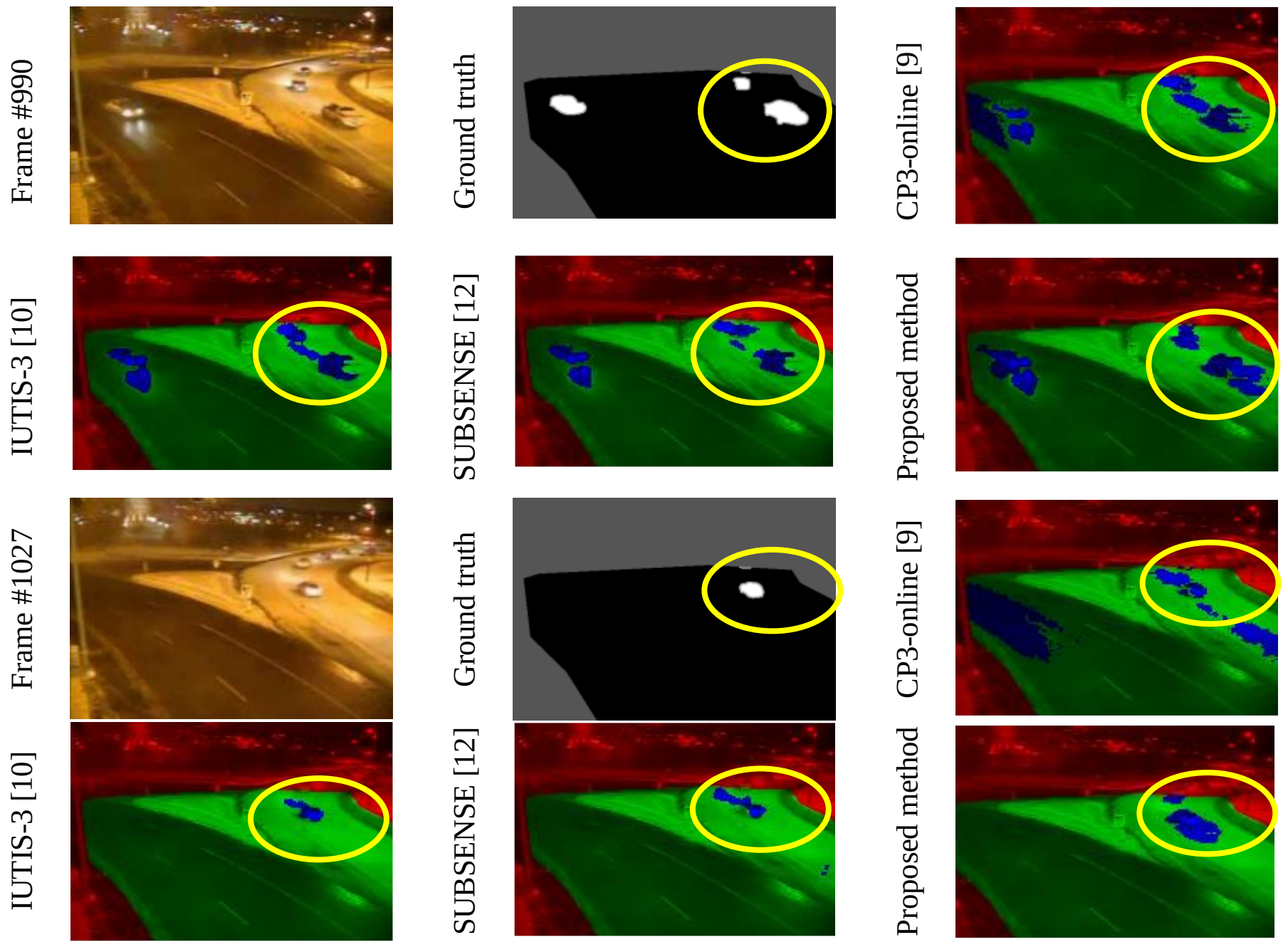

Figure 6: Comparison of the occlusion of the moving object detection for the proposed method and the previous approaches

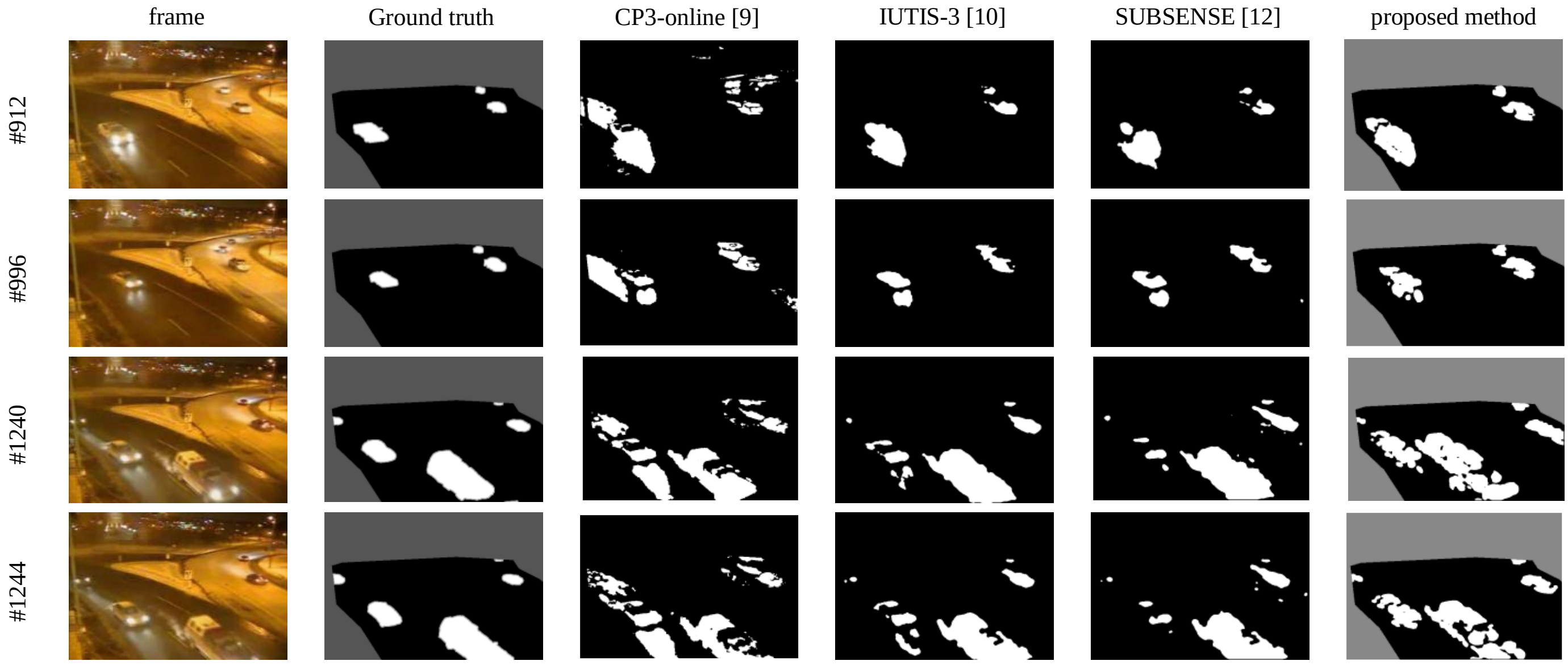

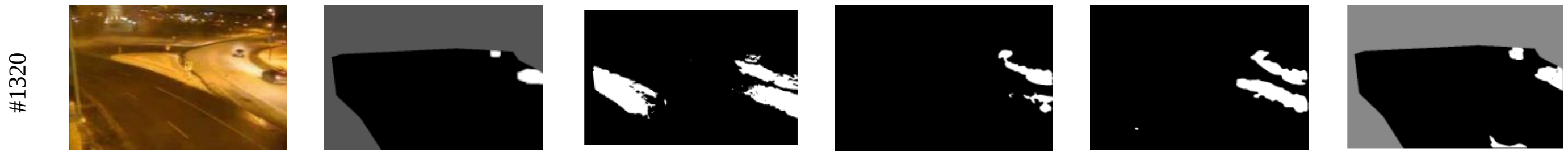

Fig.7. Qualitative comparison of the segmentation results with different approaches for various frames of '*winterStreet'* sequence of the CD.net 2014 dataset

Moreover, Figure 9 illustrates another qualitative comparison of our proposed method with CP3-online [9], SUBSENSE [12], and AAPSA [16] for various frames of the *busyBoulvard* sequence of the CD.net 2014 dataset. As these results indicate, our proposed method can overcome the illumination variations, occlusion more robustly.

Figure 10 illustrates a qualitative comparison of our proposed method with CP3-online [9] and AAPSA [16] for various frames of some *Thermal* sequence of the CD.net 2014 dataset. The results indicate the appropriate ability of our proposed method in difficulties raised by camouflage.

Figure 11 illustrates the motion detection results of CP3-online [9], SUBSENSE [12], and our proposed method respectively for various frames of the '*blizzard', 'streatlight', and 'parking'* sequences in the CD.net 2014 dataset. As the results indicate, the proposed method can segment far and tiny moving objects perfectly.

In Figures 7, 8, 9, and 11, the gray segments in the last column indicate the mask which is extracted from the ground truths and multiplied to the results of the proposed method. This mask is applied for the results of the other methods by multiplying zero for this region to the segmentation results.

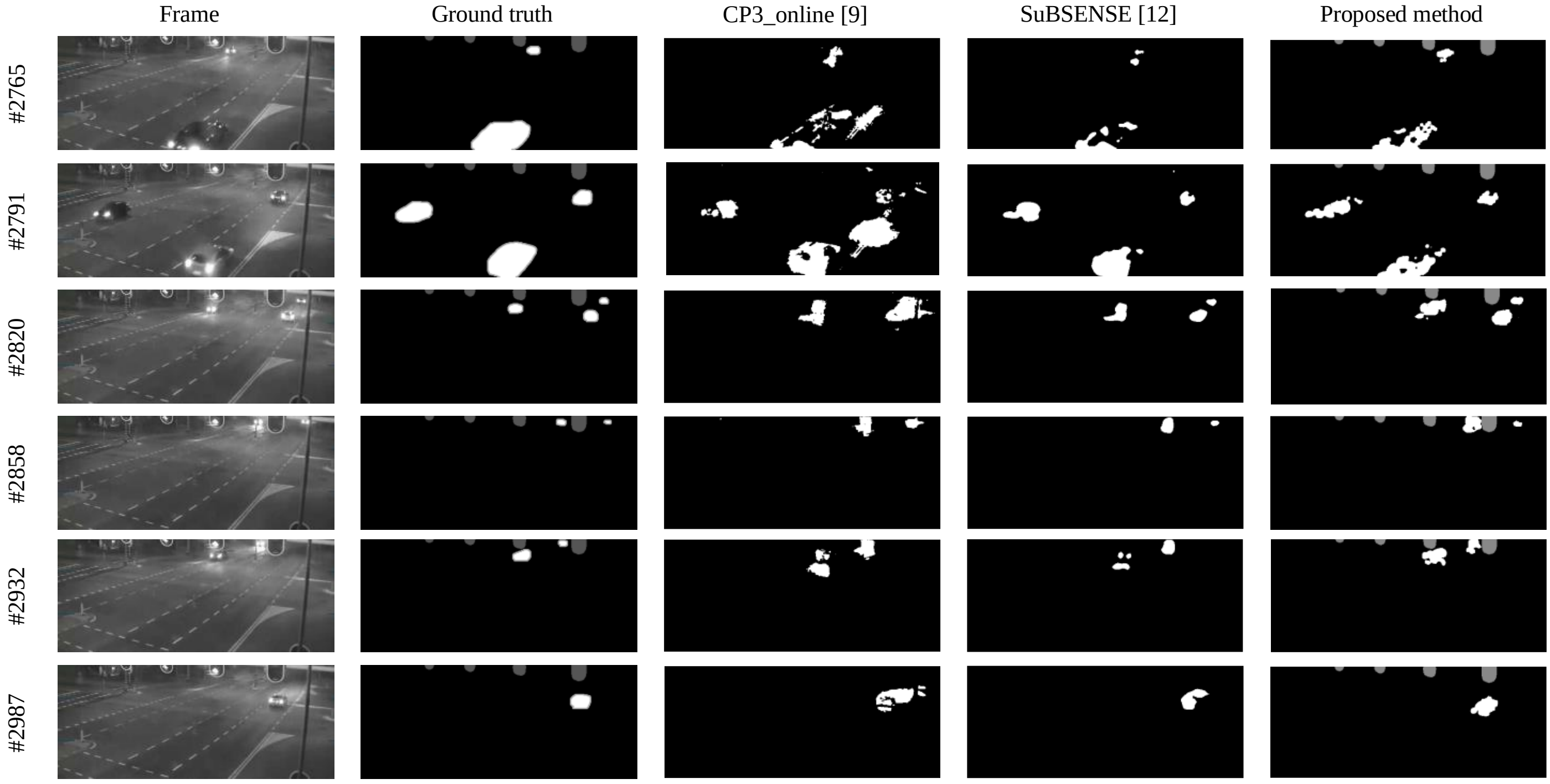

Fig.8. Qualitative comparison of the segmentation results with different approaches for various frames of '*streetCornerAtNight'* sequence of the CD.net 2014.

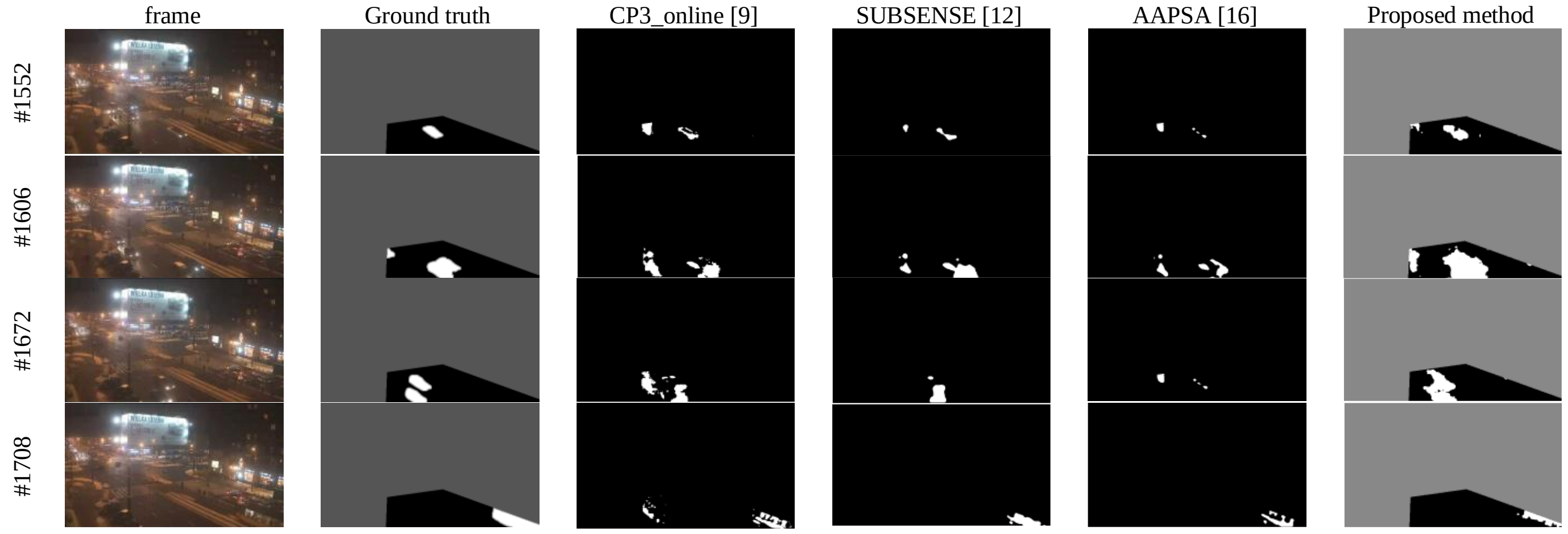

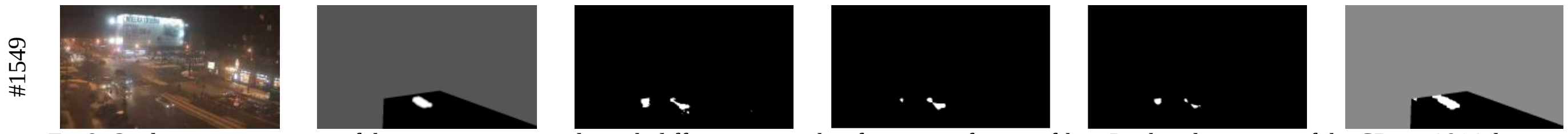

Fig.9. Qualitative comparison of the segmentation results with different approaches for various frames of *busyBoulvard* sequence of the CD.net 2014 dataset

More comparisons between the results of the proposed method and the previous approaches, without multiplying the masks of the ground truths, are available at http://dspl.ce.sharif.edu/motiondetector.html.

Furthermore, Figure 12 indicates a qualitative comparison of the motion detection results with CP3-online [9], for various frames of *busStation* sequence of the CD.net 2014 dataset.

Figure 13 indicates another qualitative comparison of the motion detection results with MDT [33] for some frames of the sequences of the UCSD pedestrian dataset. In this figure, the green segments represent motions and the yellow segments represent none-motion segments. Results indicate the noticeable efficiency improvement of our proposed method.

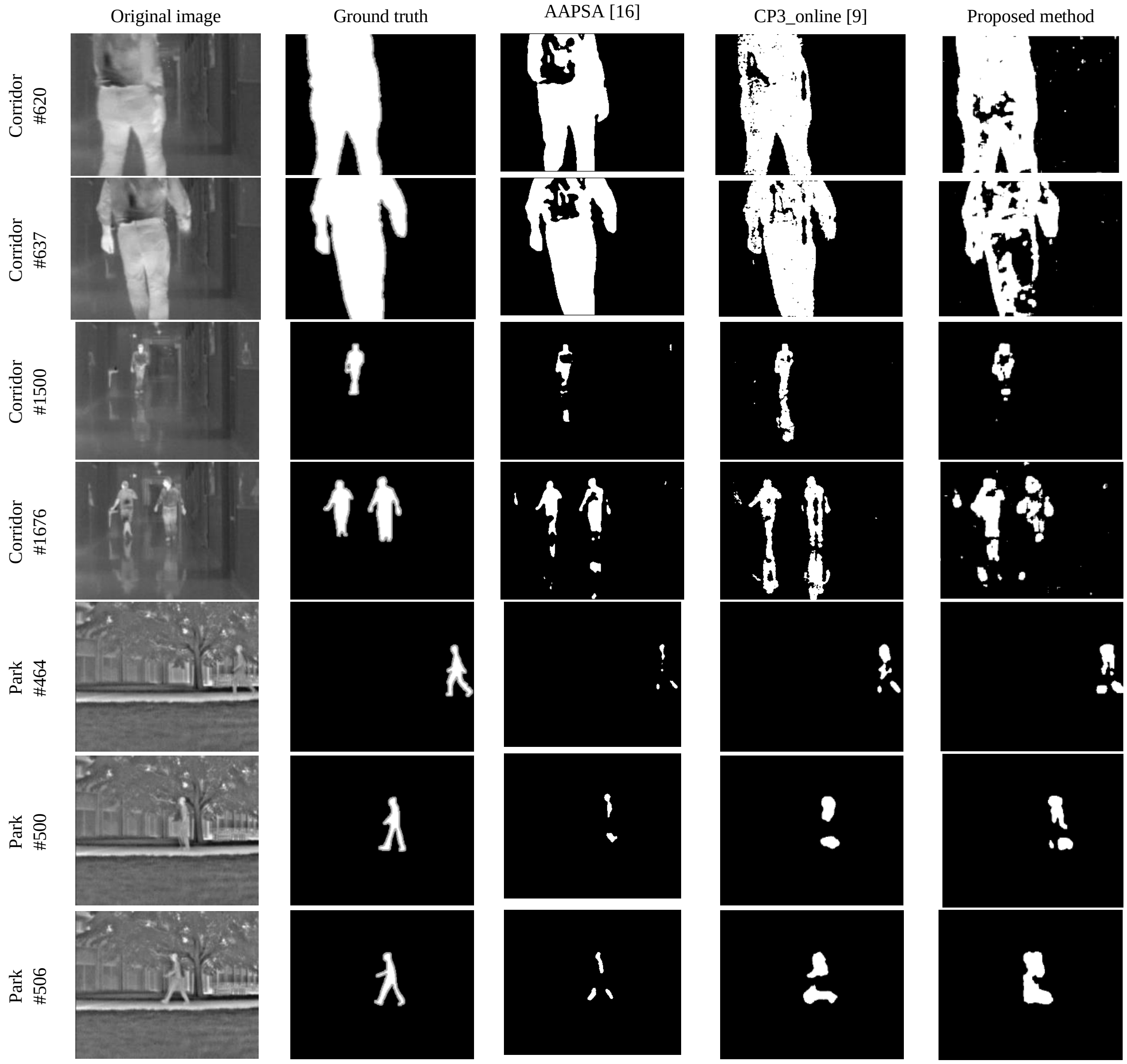

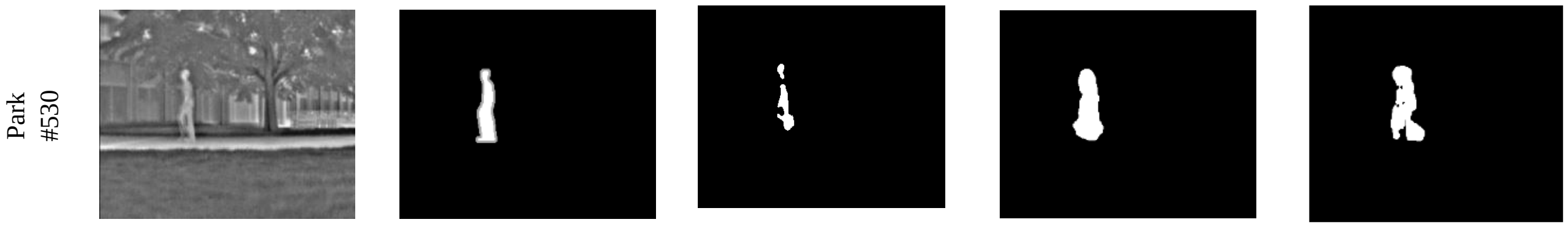

Fig.10. Qualitative comparison of the segmentation results with different approaches for various frames of some '*Thermal*' sequences of the CD.net 2014 dataset

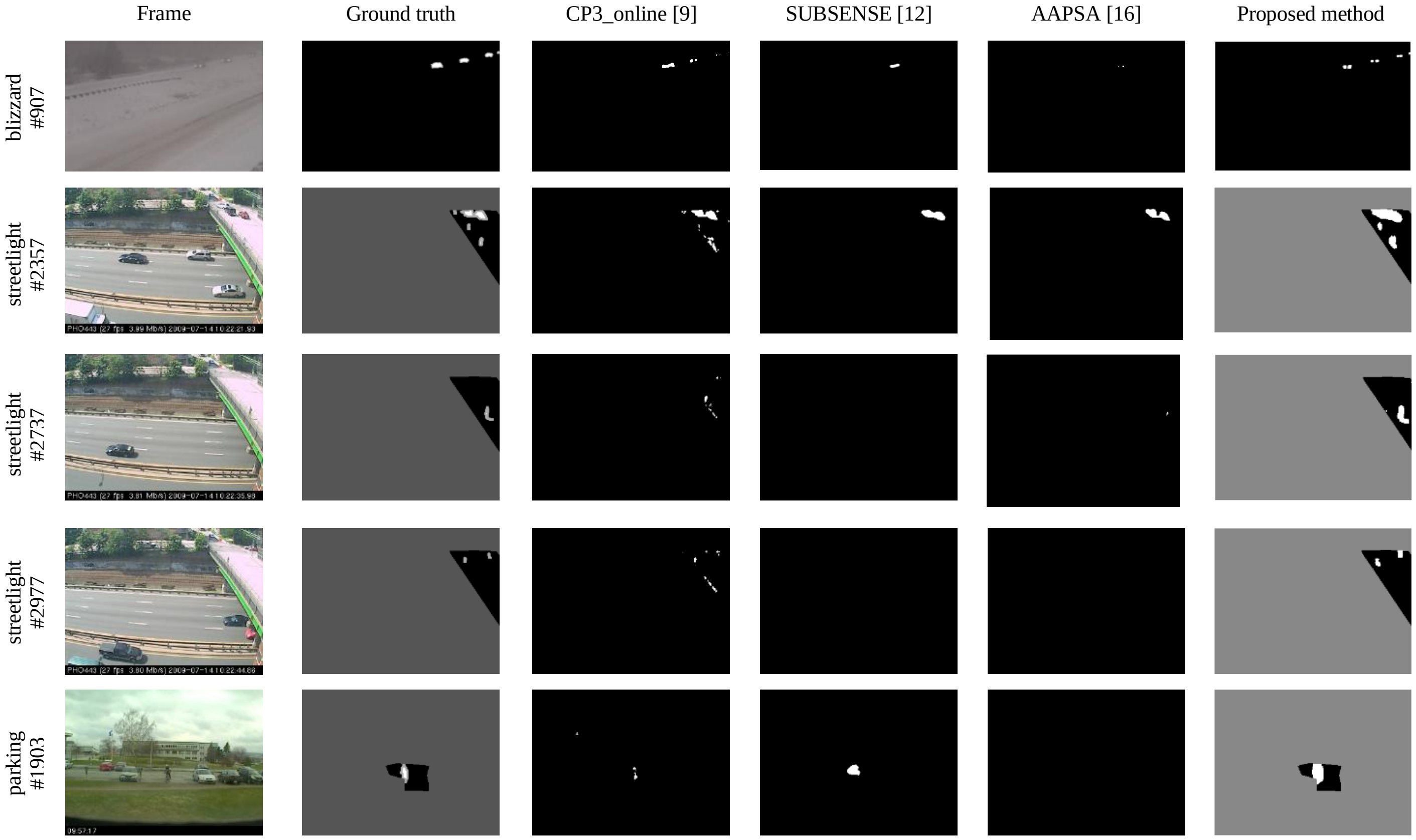

Fig.11. Qualitative comparison of the segmentation results with different approaches for some frames of *intermittentObjectMotion* of the sequences of the CD.net 2014 dataset

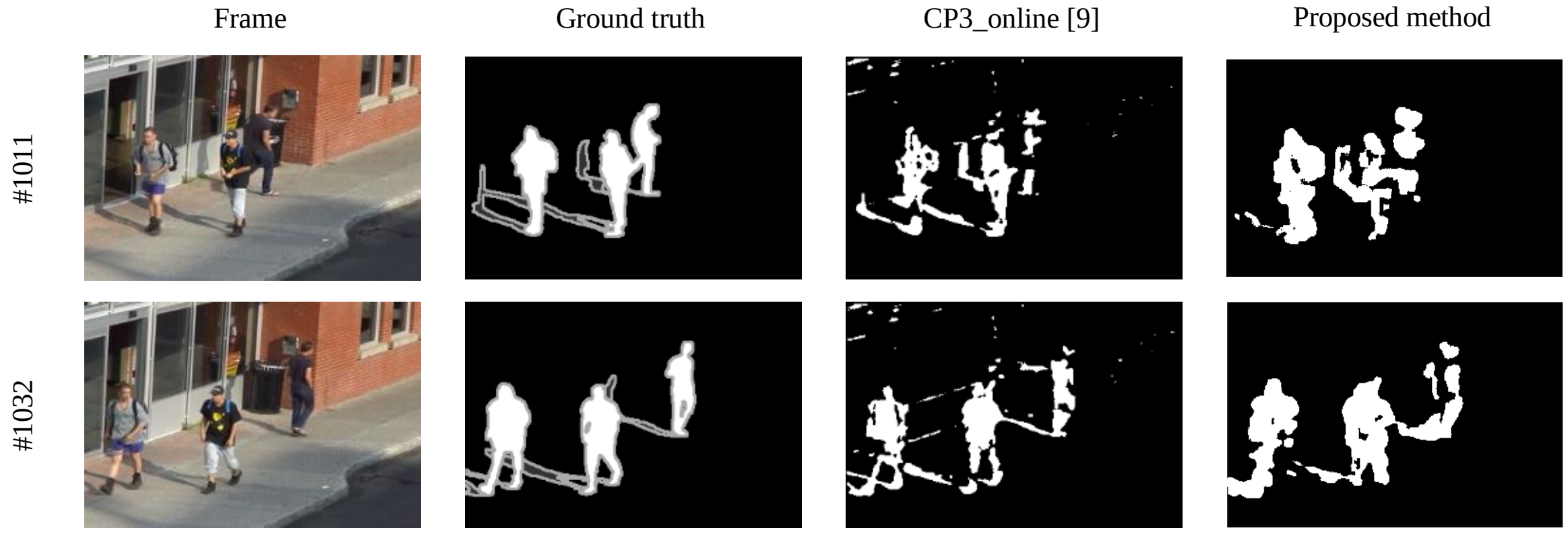

#1084

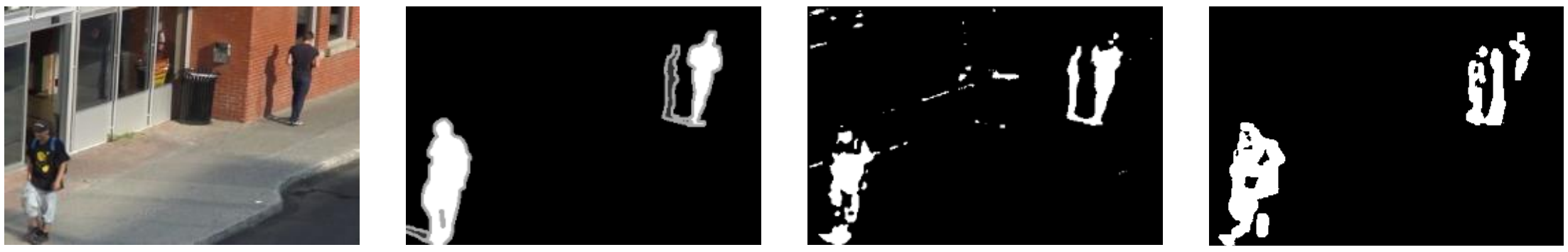

Figure 12: Qualitative comparison of the motion detection results with different approaches for various frames of *busStation* sequence of the CD.net 2014 dataset

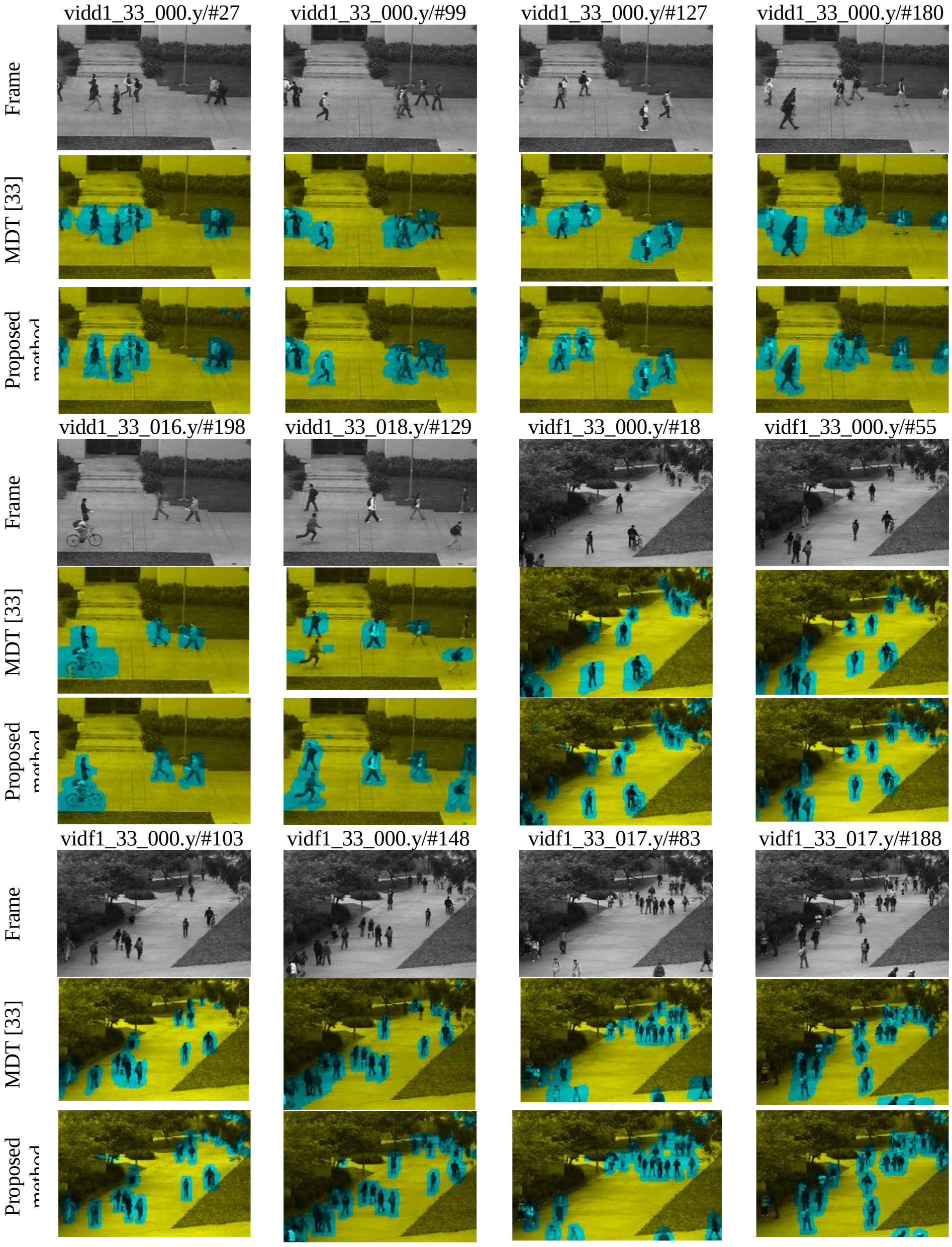

Figure 13: Qualitative comparison of the motion detection results with MDT [33] for some frames of the sequences of the UCSD pedestrian dataset, (static segments colored yellow and motion segments colored green)

Figure 14 illustrates a qualitative comparison of the motion detection results of the proposed method with CP3-online [9] for burst motion. This figure contains a cathode ray tube and a waving tree displayer respectively. As the results show, our proposed method can overcome the problem of burst motion detection properly. Finally, Figure 15 illustrates the motion detection results of our proposed method on various frames of sequences in Dyntex dataset. In this figure, the segmentation results indicate the static and dynamic regions by two different colors. The sequences contain dynamic textures like smoke, fire, and waving water. And burst motions contain disk driver and washing machine. The results indicate that our proposed method can be used for dynamic texture segmentation in video sequences.

### B. *Quantitative Mesurments*

For quantitative comparison, various evaluation metrics contain *Recall* (*Re*); *Specificity* (*Sp*); *False Positive Rate* (*FPR*); *False Negative Rate* (*FNR*); *Percentage of Wrong Classifications* (*PWC*); *F-measure*, and *Precision* will be used. Equations (6)-(12) demonstrate the quantitative measures in terms of *True Positive* (*TP*), *True Negative* (*TN*), *False Positive* (*FP*), and *False Negative* (*FN*).

*Recall* which is defined as:

$$\mathrm{Re} = \frac{TP}{TP+FN}, \tag{8}$$

can be seen as the completeness of the moving object. *Specificity* can be seen as the completeness of background and is defined as:

$$Sp = \frac{TN}{TN + FP}. \tag{9}$$

*False Positive Rate* is the rate of the background which is detected as the foreground incorrectly, and *False Negative Rate* - the rate of the foreground which is detected as the background incorrectly, which are defined as the equations (10) and (11) respectively.

$$FPR = \frac{FP}{FP + TN} \tag{10}$$

$$FNR = \frac{FN}{TP + FN} \tag{11}$$

*Percentage of Wrong Classifications* is the percentage of the foreground and background which is detected incorrectly and is defined as:

$$PWC = \frac{100*(FN + FP)}{(TP + FN + FP + TN)}. \tag{12}$$

Finally, *F-measure* is a weighted harmonic mean of the *Precision* and *Recall* and is defined as:

$$F - measure = \frac{2.\mathrm{Pr}ecision \times \mathrm{Re}}{\mathrm{Pr}ecision + \mathrm{Re}}, \tag{13}$$

in which *Precision* is defined as:

$$\mathrm{Pr}ecision = \frac{TP}{TP + FP}. \tag{14}$$

For the aforementioned measures, zero is the best value for *FPR*, *FNR*, and *PWC*, while one is the most value for *Recall*, *Specificity*, and *F-measure*.

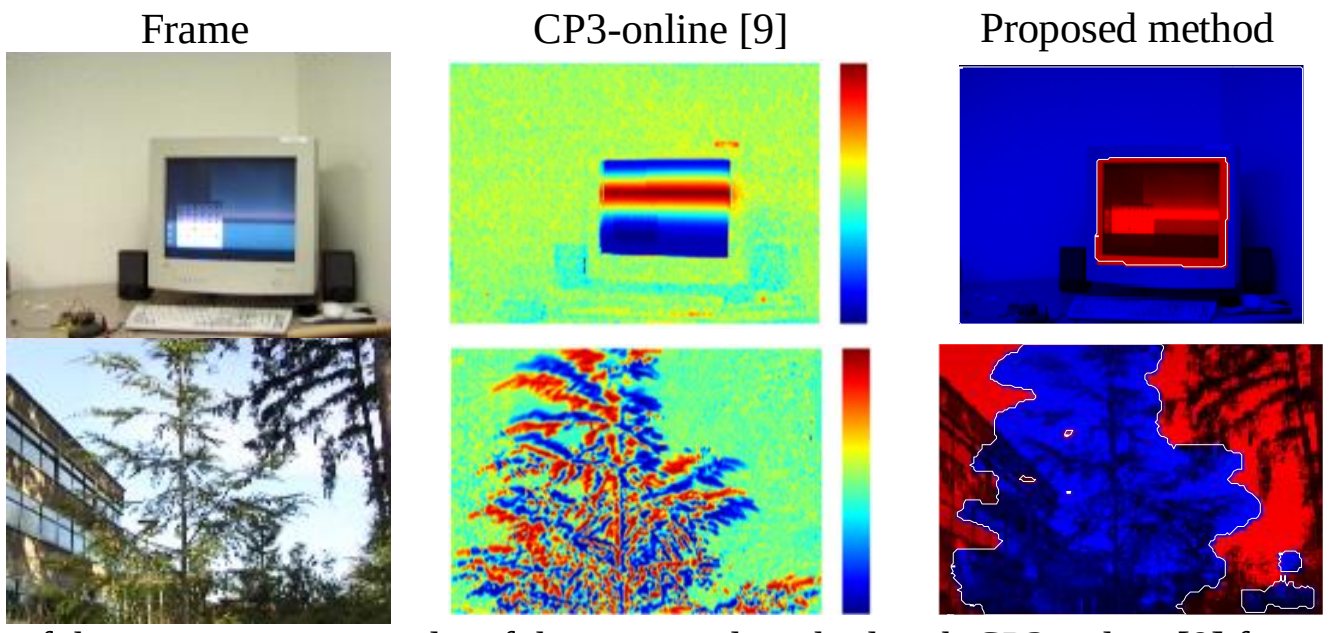

Fig.14. Qualitative comparison of the segmentation results of the proposed method with CP3-online [9] for two frames of the *Wallflower* dataset

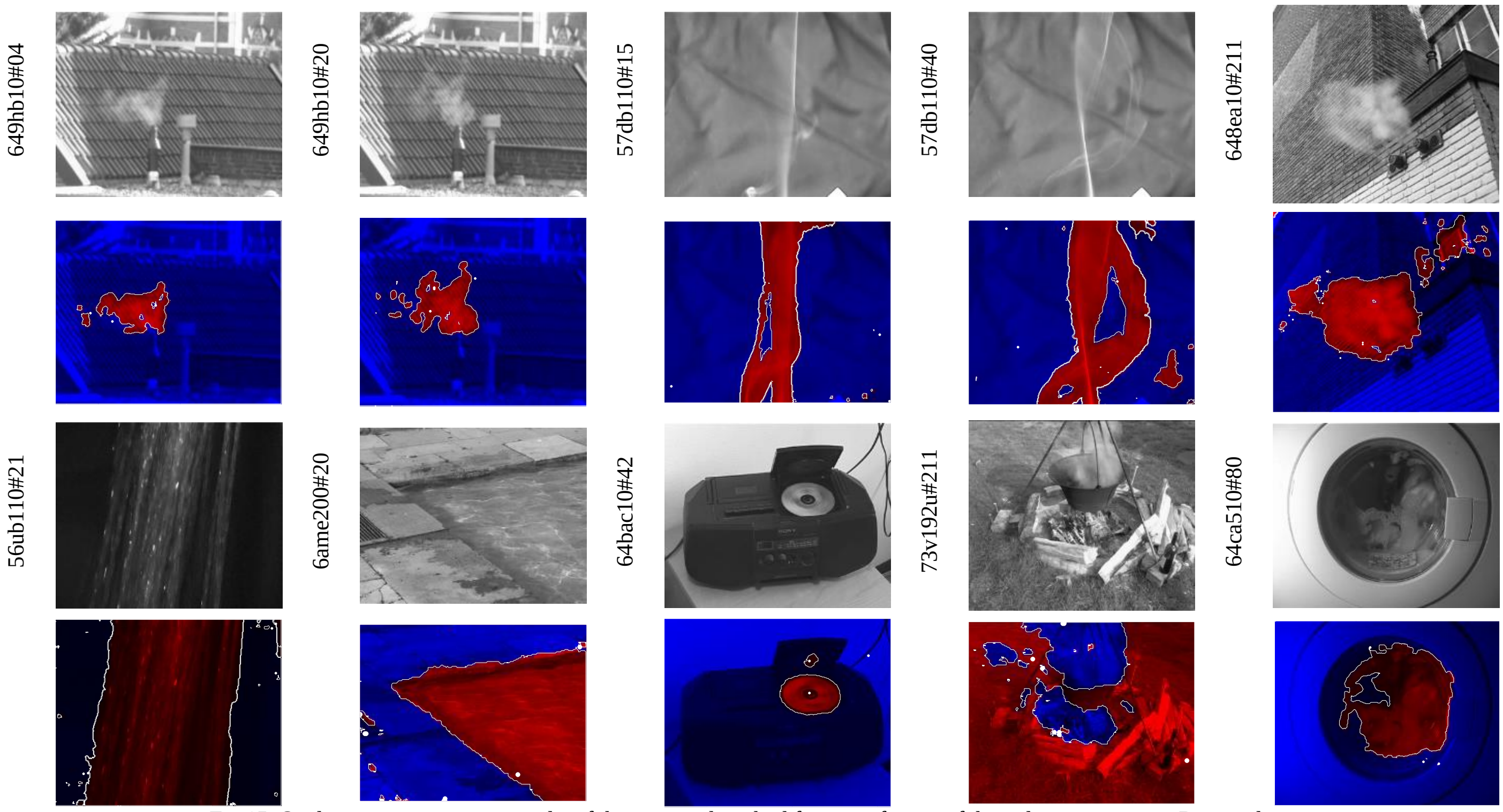

Fig.15. Qualitative segmentation results of the proposed method for some frames of the video sequences in Dyntex dataset

### C. Quantitative Comparison

As mentioned before, the most important parameter of the model is the cubic patch size for computing the wavelet coefficients. In this section, we evaluate the proposed algorithms for different cubic patch sizes which illustrated in table 1. In this table the scale decomposition levels are shown. In this section these different patch sizes are used for examining the proposed method.

Table 2 illustrates the average values of the different measures containing *Re*, *SP*, *FPR*, *FNR*, *PWC*, *Precision*, *F-measure* and the average of computational time in seconds taken by the proposed approach for the volumes with different scales and sizes, for eighteen video sequences of three different categories of CD.net 2014 dataset. The video sequences are *highway* (with frames size: 320×240), *parking* (320×240), *streetlight* (320×240), *abandonedBox* (432×288), *winterDriveway* (320×240), *tramstop* (432×288), *sofa* (320×240), *park* (352×288), *lakeSide* (320×240), *corridor* (320×240), *diningRoom* (320×240), *library* (320×240), *bridgeEntry* (630×430), *busyBoulvard* (640×364), *fluidHighway* (700×450), *streetCornerAtNight* (595×245), *tramStation* (480×295), *winterStreet* (624×420) belong to different categories of CD.net 2014 dataset contain *Nightvideos*, *Thermal*, and *intermittentObjectMotion* and *Baseline*. The approach is implemented in MATLAB 8.3 running on an Intel 3.30 Giga-Hz CPU system with 4.00 Giga-Bytes memories. As mentioned before, zero is the best value for *FPR*, *FNR*, and *PWC* measures and one is the most value for *Recall*, *Specificity*, and *F-measure*. Results indicate that the patch size 4×4×4 produces to the best results. Also, the computational time for the best patch size is 304.80 seconds.

Table 3 illustrates a quantitative comparison with various approaches containing CP3-online [9], IUTIS-3 [10], and

Table 1. The sizes of the cubic patches and their decomposition scales which are used for experimental results

| Patch sizes | Scale | Decomposition levels |
|---|---|---|
| 2×2×2 | 1 | 1×1×1 |
| 2×2×4 | 1 | 1×1×2 |
| 2×2×6 | 1 | 1×1×3 |
| 2×2×8 | 1 | 1×1×4 |
| 4×4×2 | 1 | 2×2×1 |
| 4×4×4 | 2 | 2×2×2→1×1×1 |
| 4×4×6 | 2 | 2×2×3→1×1×1 |
| 4×4×8 | 2 | 2×2×4→1×1×2 |
| 8×8×2 | 1 | 4×4×1 |
| 8×8×4 | 2 | 4×4×2→2×2×1 |
| 8×8×6 | 2 | 4×4×3→2×2×1 |
| 8×8×8 | 3 | 4×4×4→2×2×2→1×1×1 |

SUBSENSE [12], AAPSA [16], C-EFIC [34], GMM|Zivkovic [35], CwisarDH [4], SOBS_CF [2], and AMBER [36] for frames of the mentioned eighteen different video sequences of CD.net 2014 dataset respectively. In this experiment, the cubic patch sizes are (4×4×4) and the decomposition scale is 2. As the results show, the average values of the measures, containing *Re*, *Sp*, *FNR*, *PWC*, *Precision* and *F-measure* of our proposed method, for four different mentioned categories, are equal to 0.8209, 0.9392, 0.0608, 0.1719, 4.1139, 0.7884 and 0.7783 respectively. According to the value of *Precision* measure, these quantitative results indicate an impressive improvement compared with the previous approaches.

Figure 16-(a) illustrates a quantitative comparison of different measurements between the proposed method and nine different unsupervised approaches for video sequences of three categories of CD.net 2014 dataset contain *intermittentObjectMotion*, *Thermal* and *NightVideos*. In this figure the measurements are *RE*, *SP*, *Precision* and *F-Measure*

Table 2. The average of the different measures and computational time of different scales and patch sizes for eighteen video sequences belong to four categories of CD.net 2014 dataset

| Patch Size | Re | SP | FPR | FNR | PWC | Precision | F_measure | Computational Time (s) |
|---|---|---|---|---|---|---|---|---|
| **2×2×2** | 0.6596 | 0.9304 | 0.0696 | 0.3404 | 5.9398 | 0.6915 | 0.6355 | 156.63 |
| **4×2×2** | 0.6532 | 0.9295 | 0.0705 | 0.3468 | 7.9028 | 0.6966 | 0.6251 | 180.59 |
| **6×2×2** | 0.6588 | 0.9301 | 0.0699 | 0.3412 | 7.4811 | 0.7000 | 0.6306 | 212.81 |
| **8×2×2** | 0.6849 | 0.9311 | 0.0689 | 0.3151 | 6.9157 | 0.7097 | 0.6532 | 233.67 |
| **2×4×4** | 0.7890 | 0.9358 | 0.0642 | 0.2110 | 4.5688 | 0.7562 | 0.7504 | 278.62 |
| **4×4×4** | **0.8209** | **0.9392** | **0.0608** | **0.1791** | **4.1139** | **0.7884** | **0.7783** | 304.80 |
| **6×4×4** | 0.7938 | 0.9353 | 0.0647 | 0.2062 | 4.5466 | 0.7568 | 0.7513 | 352.21 |
| **8×4×4** | 0.7842 | 0.9318 | 0.0682 | 0.2158 | 4.8907 | 0.7190 | 0.7275 | 355.39 |
| **2×8×8** | 0.7705 | 0.9271 | 0.0729 | 0.2295 | 5.2155 | 0.6862 | 0.7036 | 556.68 |
| **4×8×8** | 0.7251 | 0.9285 | 0.0715 | 0.2749 | 5.3711 | 0.6702 | 0.6814 | 577.82 |
| **6×8×8** | 0.7012 | 0.9263 | 0.0737 | 0.2988 | 5.6828 | 0.6436 | 0.6570 | 617.46 |
| **8×8×8** | 0.6789 | 0.9231 | 0.0769 | 0.3211 | 6.0846 | 0.6153 | 0.6321 | 695.21 |

Table 3. Comparison of the average measures of different approaches for eighteen video sequences belong to four categories of CD.net 2014 dataset

| Method | RE | SP | FPR | FNR | PWC | Precision | F-measure |
|---|---|---|---|---|---|---|---|
| SuBSENSE | 0.7179 | 0.9866 | **0.0134** | 0.2821 | 3.1320 | 0.7272 | 0.6850 |
| AAPSA | 0.5136 | **0.9906** | 0.0094 | 0.4864 | 3.5021 | 0.7023 | 0.5535 |
| IUTIS-3 | 0.6933 | 0.9904 | 0.0096 | 0.3067 | **2.6992** | 0.7423 | 0.6838 |
| CP3-Online | 0.7410 | 0.9341 | 0.0659 | 0.2590 | 6.9928 | 0.5615 | 0.6021 |
| C-EFIC | 0.7889 | 0.9661 | 0.0339 | 0.2111 | 4.1081 | 0.7084 | 0.7140 |
| GMM\|Zivkovic | 0.5456 | 0.9798 | 0.0202 | 0.4544 | 4.6641 | 0.6490 | 0.5416 |
| CwisarDH | 0.5740 | 0.9904 | 0.0096 | 0.4260 | 3.3825 | 0.7098 | 0.5860 |
| SOBS_CF | 0.7110 | 0.9599 | 0.0401 | 0.2890 | 4.9629 | 0.6113 | 0.5938 |
| AMBER | 0.6997 | 0.9728 | 0.0272 | 0.3003 | 4.0272 | 0.6667 | 0.6306 |
| 3D-DWT_MD | **0.8209** | 0.9392 | 0.0608 | **0.1791** | 4.1139 | **0.7884** | **0.7783** |

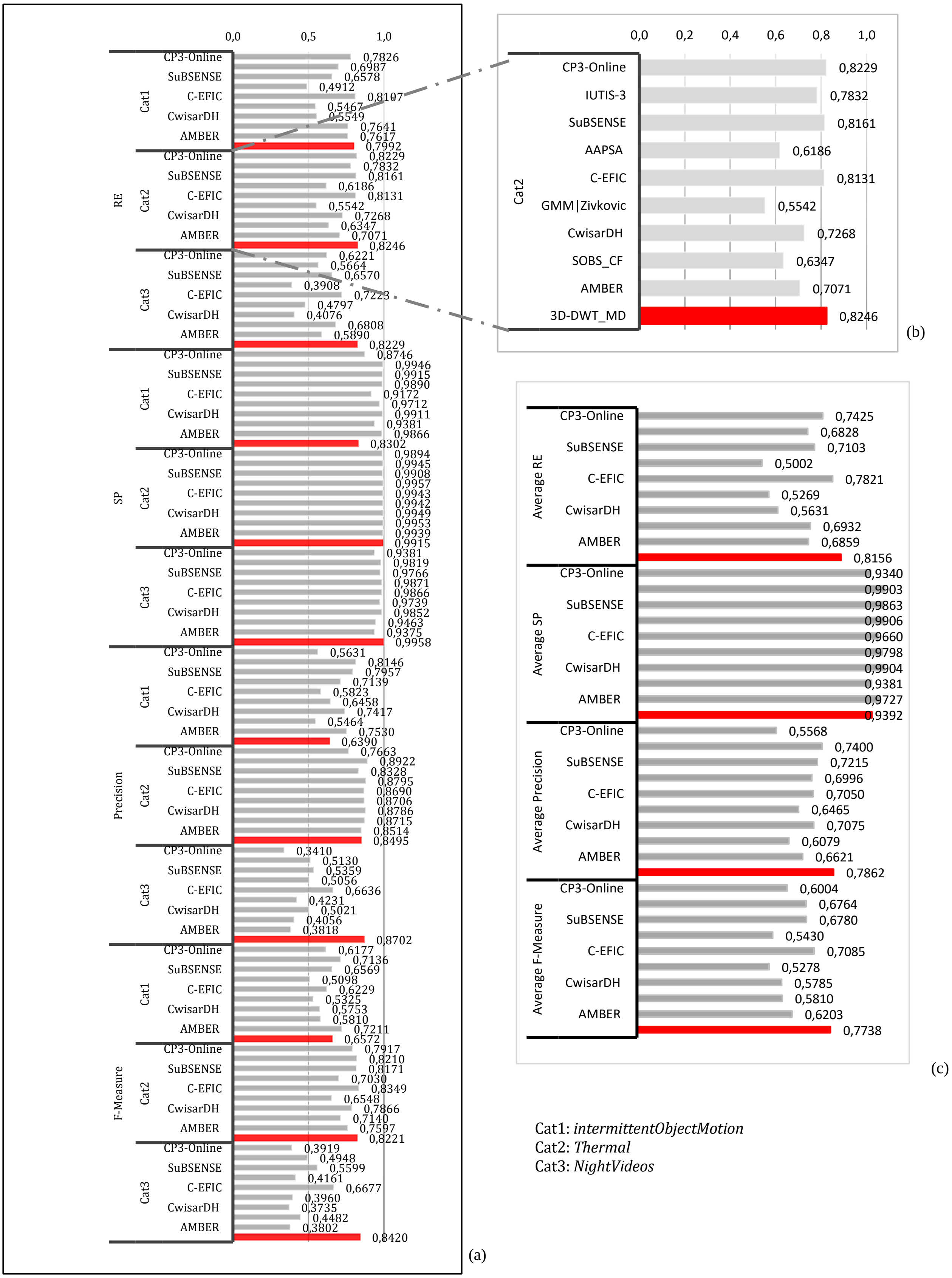

Cat1: *intermittentObjectMotion*
Cat2: *Thermal*
Cat3: *NightVideos*

Fig.16. Comparison of proposed method with nine different approaches for all the video sequences of three categories contain *intermittentObjectMotion*, *Thermal* and *NightVideos*

and the approaches are CP3-online [9], IUTIS-3 [10], and SUBSENSE [12], AAPSA [16], C-EFIC [34], GMM|Zivkovic [35], CwisarDH [4], SOBS_CF [2], and AMBER [36] and 3D-DWT_MD. In this figure, the horizontal bars in red demonstrate the measurement value for the proposed method.

Figure 16-(b) demonstrates the order of approaches in Figure 16-(a), for measuring *RE* and the second category (*Thermal*). This order is repeated for other parts of this figure intermittently. Figure 16-(c) illustrates the average of the mentioned measure for the three mentioned categories in figure 16-(a) with the same order of approaches. Comparing the measures *RE*, *Precision* and *F-Measure* in this figure indicates that the proposed method excels the other approaches significantly. Also, more details about the measure values can be find at http://dspl.ce.sharif.edu/motiondetector.html.

## VI. CONCLUSIONS

In this paper, we proposed a novel motion detection method using spatial frequency descriptors based on the three-dimensional wavelet transform and three-dimensional wavelet leader. Thanks to the ability of the frequency domain approaches in considering the holistic motion pattern information, the proposed method can effectively deal with the difficulties raised by illumination changes, camouflage, and burst physical motions. Moreover, for this valuable property of wavelet-based descriptors, the proposed method could be used for dynamic texture segmentation. Also, the proposed method had a good capability in detecting far and tiny moving objects.

In order to evaluate the performance of the proposed method, various qualitative and quantitative comparisons were done. Towards this goal, four different datasets containing CD.net 2014, Wallflower, Dyntex, and UCSD pedestrian were used. Furthermore, various evaluation metrics containing *Recall* (*Re*), *Specificity* (Sp), *False Positive Rate* (*FPR*), *False Negative Rate* (*FNR*), *Percentage of Wrong Classifications* (*PWC*), *Precision*, and *F-measure* for different video sequences, were computed. The results from these qualitative and quantitative comparisons demonstrated that our proposed approach outperforms its competitors, both in terms of motion detection and the capability of segmenting the dynamic textures properly.